\documentclass[conference]{IEEEtran}
\IEEEoverridecommandlockouts
\usepackage{times}

\usepackage[numbers]{natbib}
\usepackage{multicol}
\usepackage[bookmarks=true]{hyperref}
\usepackage{graphicx} 
\usepackage{algorithm}
\usepackage{algpseudocode}
\usepackage{amsmath} 
\usepackage{tabularx} % For X column type
\usepackage{booktabs} % For better table lines
\usepackage{rotating} % For sidewaystable
\usepackage{multicol}
\usepackage{subfigure} 

\begin{document}

% paper title
\title{Enhancing Stress-Strain Predictions with Seq2Seq and Cross-Attention based on Small Punch Test}

\author{\authorblockN{Zhengni Yang$^1$$^,$$^2$$^,$$^3$,
Rui Yang$^1$$^,$\authorrefmark{1}, Weijian Han$^2$$^,$\authorrefmark{1},
Qixin Liu$^1$}
\authorblockA{$^1$School of Advanced Technology Xi'an Jiaotong-Liverpool University, Suzhou, China, 215123} 
\authorblockA{$^2$Materials Bigdata and Applications Division Materials Academy JITRI, Suzhou, China, 215123}
\authorblockA{$^3$Department of Materials, Design and Manufacturing Engineering University of Liverpool, Liverpool, United Kingdom, L69 3BX}
\thanks{$^*$Corresponding author: Rui Yang (r.yang@xjtlu.edu.cn), Weijian Han (hanwj@mat-jitri.cn)}
\thanks{This research is partially supported by: Jiangsu Provincial Qinglan Project and Suzhou Science and Technology Programme (SYG202106).}}
\maketitle

\begin{abstract}
This paper introduces a novel deep-learning approach to predict true stress-strain curves of high-strength steels from small punch test (SPT) load-displacement data. The proposed approach uses Gramian Angular Field (GAF) to transform load-displacement sequences into images, capturing spatial-temporal features and employs a Sequence-to-Sequence (Seq2Seq) model with an LSTM-based encoder-decoder architecture, enhanced by multi-head cross-attention to improved accuracy. Experimental results demonstrate that the proposed approach achieves superior prediction accuracy, with minimum and maximum mean absolute errors of 0.15 MPa and 5.58 MPa, respectively. The proposed method offers a promising alternative to traditional experimental techniques in materials science, enhancing the accuracy and efficiency of true stress-strain relationship predictions.
\end{abstract}

\IEEEpeerreviewmaketitle

\section{Introduction}
Accurately determining the true stress-strain relationship of materials is a fundamental aspect of engineering design and structural integrity assessments \cite{Zhang2019AdvancedMaterials,yang2025machine}. Traditional methods such as tensile, compression and shear tests are costly, time-consuming, and sample-intensive \cite{Kamaya2018SmallPunchTesting, klevtsov2009method}. The small punch test (SPT) offers a viable alternative, reducing sample sizes and minimizing component damage, particularly in challenging testing scenarios \cite{cao2022determination, li2018construction,cornaggia2018inverse, janvca2016small}. Advancements in machine learning have transformed the fields of materials engineering and fault diagnosis, offering precise predictive capabilities \cite{s21237894, 10193811, 10138915, 10198368}, and the machine learning algorithms accelerate material characterization and decrease reliance on traditional methods by capturing complex data relationships \cite{HatakeyamaSato2021MachineLF, STERJOVSKI2005536,zhan2021machine, wang2023prediction}.

Among the various machine learning methods, the sequence-to-sequence (Seq2Seq) model has attracted considerable attention for its effectiveness in handling sequential data \cite{Sutskever2014SequenceToSequence}. The Seq2Seq model was initially designed for natural language tasks due to its effectiveness in handling sequential data through an encoder-decoder architecture, which predicts and decodes sequences \cite{Georgiev2017Seq2SeqASR}. The Seq2Seq model captures intricate dependencies in complex data like load-displacement and stress-strain sequences, and the encoder-decoder structure transforms input data into a context vector for accurate output generation \cite{10637749}. This ability to precisely handle sequential data highlights the potential of the Seq2Seq model as a robust tool for modeling and predicting material behaviors, offering an efficient alternative to traditional experimental approaches in materials science.

In material science, the relationship between SPT data and stress-strain data can be viewed as a translation task, akin to the mapping between source and target languages. In this study, the Seq2Seq model is applied to predict stress-strain curves from SPT data, using the load-displacement curve as the input sequence and the corresponding stress-strain curve as the output sequence. This application allows the model to learn the complex mapping between the input and output sequences, offering a novel approach for predicting stress-strain sequence data from SPT data. The Seq2Seq model provides an efficient and accurate means of predicting material behavior, reducing the reliance on extensive experimental data and offering new possibilities for rapid material property evaluation.

Integrating the Seq2Seq model into materials science highlights the transformative potential of machine learning and deep learning in predicting and understanding material behaviors. By using SPT data, the proposed approach offers a more efficient and cost-effective alternative to traditional methods, accelerating material characterization. This paper presents a novel method for predicting the true stress-strain relationship of high-strength steels using the Seq2Seq structure, advancing the application of machine learning in material behavior prediction. The main contributions of this study are as follows:
\begin{enumerate} 
  \item This study introduces a novel deep-learning framework that employs a Seq2Seq model with multi-head cross-attention to improve prediction accuracy and efficiency compared to traditional methods; 
  \item This study proposes a feature extraction method using multi-scale convolution layers to capture features from the original sequence and Gramian Angular Field (GAF) images, further enhancing the model's predictive accuracy.
\end{enumerate}
The remainder of this paper is organized as follows: Section \ref{Methodology} details the proposed methodology, Section \ref{Experimental Results and Discussions} presents the experimental results and discussions, and Section \ref{Conclusion} concludes the study and suggests directions for future research.

\section{Methodology}
\label{Methodology}
\begin{figure}[htp]
    \centering
    \includegraphics[width=1\linewidth]{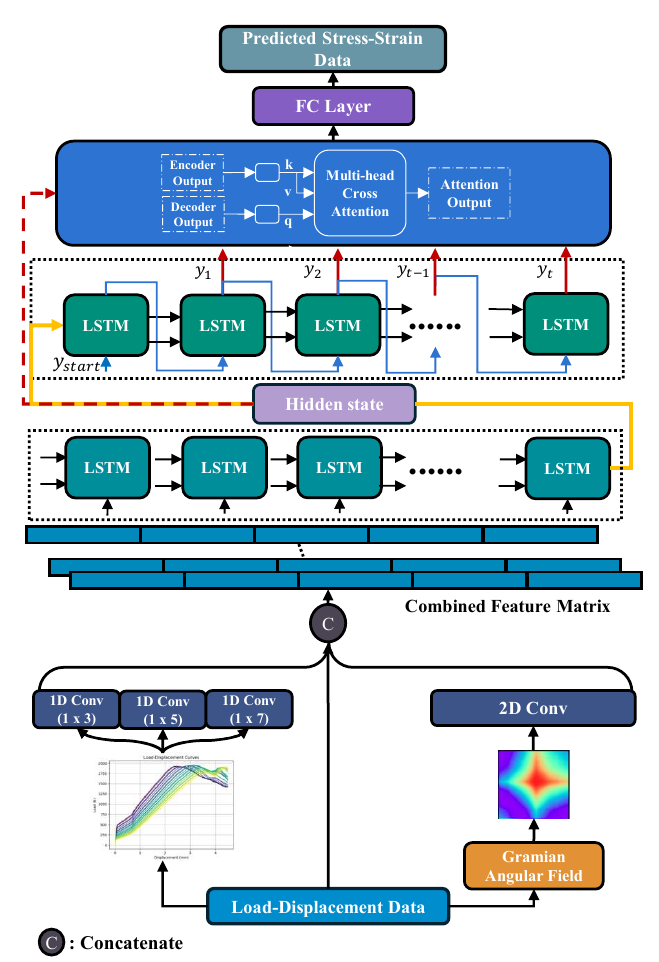}
    \caption{Structure of Proposed Approach}
    \label{fig: structure of proposed approach}
\end{figure}

This study proposes a novel method to predict true stress-strain curves from SPT data. The proposed approach utilizes SPT data as input and employs a Seq2Seq model combined with a multi-head cross-attention mechanism to accurately predict stress-strain curves for high-strength steel materials. The architecture of the proposed method is illustrated in Fig. \ref{fig: structure of proposed approach}. This section provides a detailed description of the data generation process, the feature extraction algorithms, and the structure of the proposed model employed in this approach.

\subsection{Data Generation}

This study generated training and testing data for high-strength steel using LS-DYNA software to simulate load-displacement and stress-strain curves. The finite element method (FEM) modeled the SPT and tensile tests, accurately replicating sample geometry and loading conditions. Simulations covered materials with varying properties and thicknesses from 1 to 4 mm, using Young’s modulus of 70,000 MPa and a Poisson’s ratio of 0.35.

\begin{figure}[!t]
	\centering
        \subfigure[Progressive Stages of Punch Impact on the Simulation Model]{
		\begin{minipage}[b]{0.45\textwidth}
			\includegraphics[width=1\textwidth]{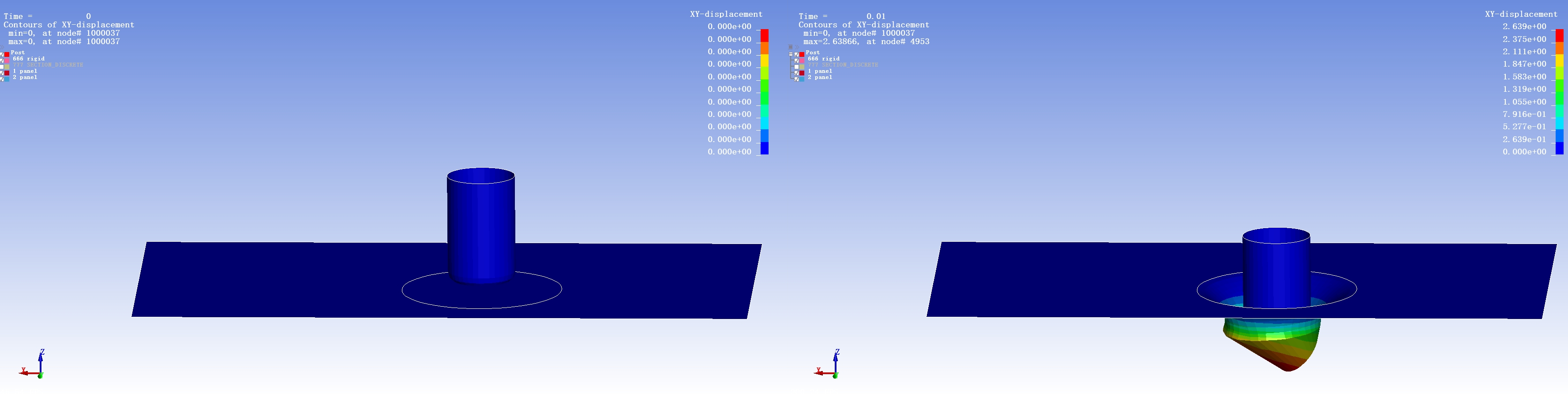}
		\end{minipage}
		\label{fig: Progressive Stages of Punch Impact on the Simulation Model}
        }
        \subfigure[Progressive Stages of Tensile Test on the Simulation Model]{
        \begin{minipage}[b]{0.45\textwidth}
            \includegraphics[width=1\textwidth]{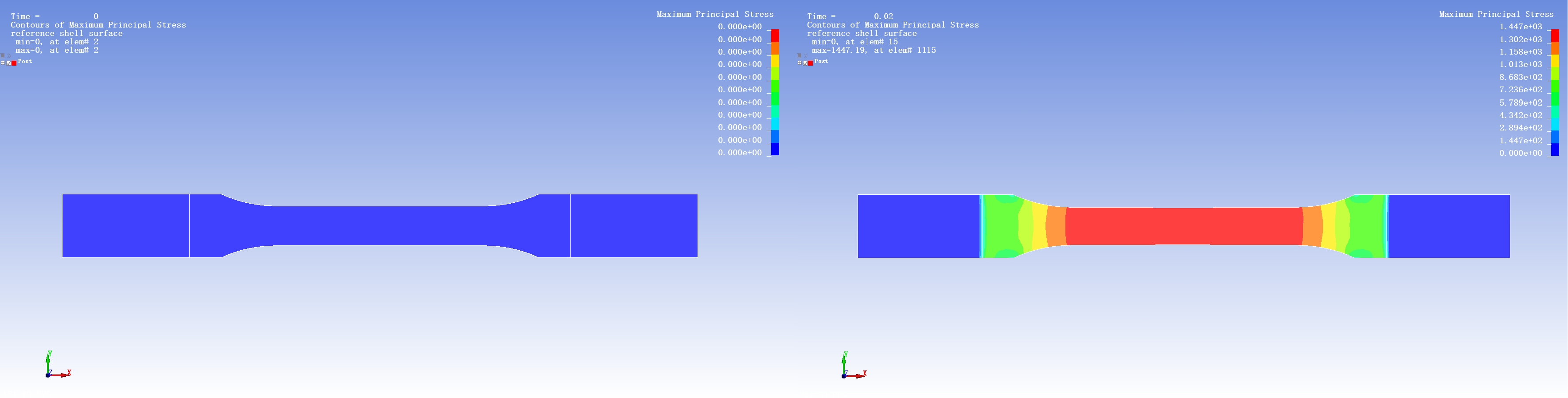}
        \end{minipage}
        \label{fig: Progressive Stages of Tensile Test on the Simulation Model}
        }
	\caption{FEM Configurations for Small Punch Test and Tensile Test}
	\label{Progressive Stages}
\end{figure}

For the SPT simulation, the load was set to 20 kN, and the puncher was pressed into the specimen at a constant displacement rate of 10 mm/min, with load and displacement data recorded throughout the process. For tensile testing, the stretching rate was set to 7.2 mm/min, and stress-strain data were collected. This setup enabled the generation of 5,000 high-strength steel samples with varying mechanical behaviors, forming the basis for the training and testing datasets used in this study. Fig.~\ref{Progressive Stages} illustrates the data generation process using LS-DYNA: Fig.~\ref{fig: Progressive Stages of Punch Impact on the Simulation Model} presents a schematic of the simulation mechanism, showing the stages of punch impact on the model, while the Fig.~\ref{fig: Progressive Stages of Tensile Test on the Simulation Model} depicts the simulation setup for tensile testing. Overall, in this study, 5000 high-strength steel samples with varying mechanical behaviors are generated for training and testing.

\subsection{Feature Extraction}
\label{Feature Extraction}

\begin{algorithm}
\caption{Training Procedure for the Proposed Model}
\label{algorithm1}
\begin{algorithmic}[1]
\State \textbf{Input:} Load-displacement data $\mathcal{D}$
\State \textbf{Output:} Predicted stress-strain data $\mathcal{P}$
\State \textbf{Preprocess Data:}
    \State \quad Compute Gramian Angular Field $\mathcal{G}$ using \eqref{eq1} and \eqref{eq2}
    \State \quad Extract 2D features $\mathcal{F}_{2D}$ from $\mathcal{G}$ using 2D Conv layers
    \State \quad Extract 1D features $\mathcal{F}_{1D}$ from $\mathcal{D}$ using 1D Conv layers
    \State \quad Obtain feature matrix $\mathcal{M}$ using \eqref{eq3}
\State \textbf{Train Seq2Seq Model:}
    \State Initialize encoder $f_{\text{encoder}}$, decoder $f_{\text{decoder}}$, and cross-attention $\mathcal{A}$
    \State Set optimizer to Adam and loss function
    \For{each epoch $\epsilon = 1, \dots, \text{num\_epochs}$}
        \For{each batch $\mathcal{B}$ in $\mathcal{M}$}
            \State Prepare input sequence $\mathcal{X}$ and target sequence $\mathcal{Y}$
            \State Encode $\mathcal{X}$ to obtain hidden states $\mathcal{H}$ using \eqref{eq4}
            \State Initialize batch loss $\mathcal{L} \gets 0$
            \For{each time step $t = 1, \dots, \text{length of } \mathcal{Y}$}
                \State Generate decoder output $\mathcal{O}_t$ using \eqref{eq5}
                \State Compute context vector $\mathcal{C}_t$ using \eqref{eq6}
                \State Compute prediction $\mathcal{P}_t$ from $\mathcal{C}_t$
                \State $\mathcal{L} \gets \mathcal{L} + \text{criterion}(\mathcal{P}_t, \mathcal{Y}_t)$
            \EndFor
            \State Backpropagate $\mathcal{L}$ and update parameters
        \EndFor
    \EndFor
\end{algorithmic}
\end{algorithm}

In this study, the GAF method \cite{WANG201568} transforms load-displacement data into sequence images, enhancing feature extraction for deep learning models. By mapping data points to polar coordinates, GAF preserves sequence order and magnitude, aiding in effective analysis. Specifically, the angles \( \theta_i \) and \( \theta_j \) are computed for each pair of points \( x_i \) and \( x_j \) in the sequence as:  
\begin{equation}
\begin{aligned}
\theta_i = \arccos\left(\frac{x_i - \min(\mathcal{D})}{\max(\mathcal{D}) - \min(\mathcal{D})}\right)\\
\theta_j = \arccos\left(\frac{x_j - \min(\mathcal{D})}{\max(\mathcal{D}) - \min(\mathcal{D})}\right)
\end{aligned}
\label{eq1}
\end{equation}
The GAF image is then constructed by calculating the cosine of the angular sum for each pair of points, forming a symmetric matrix defined as:
\begin{equation}
    \mathcal{G}_{ij} = \cos(\theta_i + \theta_j)
\label{eq2}
\end{equation}
Where $\mathcal{G}_{ij}$ represents the element at the \( i \)-th row and \( j \)-th column of the symmetric matrix, while \( \theta_i \) and \( \theta_j \) are the angles corresponding to the \( i \)-th and \( j \)-th points in the sequence, respectively. The GAF transformation retains both the temporal dependencies and intrinsic correlations within the data, capturing the complex, nonlinear relationships that are characteristic of load-displacement sequences in high-strength steel.

The GAF images are processed by a two-dimensional convolutional neural network (2D Conv) to extract features $\mathcal{F}_{2D}$, capturing correlations in load-displacement data.  A one-dimensional convolutional network (1D Conv) applies kernel sizes of 3, 5, and 7 to extract temporal feature $\mathcal{F}_{1D}$ from the original sequence. Combining $\mathcal{F}_{1D}$ and $\mathcal{F}_{2D}$ produces a comprehensive feature representation matrix.

To construct a complete feature matrix for each sample, the extracted features from both the 1D Conv and the 2D Conv are combined with the original SPT data using a residual connection, as shown in the following equation:
\begin{equation}
\mathcal{M} = \text{Concat}(\mathcal{D}, \mathcal{F}_{1D}, \mathcal{F}_{2D})
\label{eq3}
\end{equation}
This concatenated feature matrix creates a robust and unified representation of each sample, integrating diverse aspects of the data. The comprehensive feature matrix serves as the foundation for the subsequent stages of the predictive modeling process, enhancing the model's capacity to learn complex patterns and dependencies required for accurate prediction of stress-strain curves in high-strength steels. The feature extraction process is detailed in Algorithm~\ref{algorithm1}.

\subsection{LSTM-Based Seq2Seq Model with Multi-Head cross-attention Module}
\label{LSTM}
The proposed method utilizes an LSTM-based Seq2Seq model enhanced by the multi-head cross-attention mechanism to predict stress-strain curves from the feature matrix. The multi-head cross-attention aligns each step of the decoder output with the encoder’s hidden state, capturing the sequence patterns within the data. By using the cross-attention algorithm, the model continuously refines its predictions by referencing the encoder’s hidden state, as shown in Algorithm~\ref{algorithm1}.

The LSTM encoder-decoder architecture processes the feature matrix ($\mathcal{M}$). The encoder consists of five layers designed to capture complex, nonlinear relationships in sequential data. Each LSTM cell has a hidden size of 128 and uses dropout to reduce overfitting, and the decoder has the same structure. The encoder's output, stored in the hidden state, preserves temporal dependencies for the decoder's predictions. For the input sequence \(\mathcal{X}\), the procedures of the encoder are as follows:
\begin{equation}
    \mathcal{H}_t = f_{\text{encoder}}(\mathcal{X}_t, \mathcal{H}_{t-1}), \quad \forall t
\label{eq4}
\end{equation}
where \( \mathcal{H}_t \) is the hidden state at time \( t \), and \( f_{\text{encoder}} \) is the LSTM encoder function.

The decoder, also consisting of five LSTM layers with a hidden size of 128 and dropout, utilizes the hidden state generated by the encoder to predict the target stress-strain data. Initialized with the encoder’s hidden state, the decoder preserves the temporal dependencies learned by the encoder throughout the sequence generation process:
\begin{equation}
    \mathcal{O}_t = f_{\text{decoder}}(\mathcal{O}_{t-1}, \mathcal{S}_{t-1})
\label{eq5}
\end{equation}
where \(\mathcal{O}_t\) is the output of the decoder at time \( t \), and \(\mathcal{O}_{t-1}\) and \(\mathcal{S}_{t-1}\) are the output and hidden state from time \( t-1 \), respectively.

To enhance prediction accuracy, the model employs a multi-head cross-attention mechanism to align the decoder with the encoder's hidden states at each step. By dividing attention into multiple heads, the model simultaneously focuses on different parts of the encoder's output, improving robustness and expressiveness \cite{gheini-etal-2021-cross}. This mechanism preserves temporal relationships in the original sequence, refining predictions and ensuring consistency with learned sequence patterns. The cross-attention mechanism is defined as:
\begin{equation}
\begin{aligned}
      \alpha_{t,i} = \frac{\exp(\mathcal{H}_i^\top \mathcal{O}_t)}{\sum_{j} \exp(\mathcal{H}_j^\top \mathcal{O}_t)}, \quad \mathcal{C}_t = \sum_{i} \alpha_{t,i} \mathcal{H}_i  
\end{aligned}
\label{eq6}
\end{equation}
where \( \alpha_{t,i} \) represents the attention weights, and \( \mathcal{C}_t \) is the context vector. The output of the cross-attention mechanism is then passed through a fully connected layer to generate the predicted stress-strain curve. By combining the LSTM-based Seq2Seq architecture with the multi-head cross-attention mechanism, the model effectively learns complex patterns and achieves precise predictions of material behavior.

In summary, this study presents a novel deep learning-based approach for predicting true stress-strain curves from SPT load-displacement data. The methodology begins with applying the GAF transformation, which converts SPT data into GAF images to enable spatial feature extraction using a 2D Conv. Simultaneously, a 1D Conv with multiple kernel sizes captures temporal patterns from the original sequence. The extracted features are combined with the original sequence data through residual connections, forming a comprehensive feature matrix. The feature matrix is processed by a Seq2Seq model to effectively capture complex sequential dependencies, while a multi-head cross-attention mechanism aligns each decoder prediction with the encoder's hidden states, enhancing accuracy. This integrated approach offers a robust framework for accurately modeling stress-strain relationships in high-strength steels.

\section{Experimental Results and Discussions}
\label{Experimental Results and Discussions}

\begin{figure}[!t]
	\centering
        \subfigure[Load-Displacement Curves of Training Samples]{
		\begin{minipage}[b]{0.22\textwidth}
			\includegraphics[width=1\textwidth]{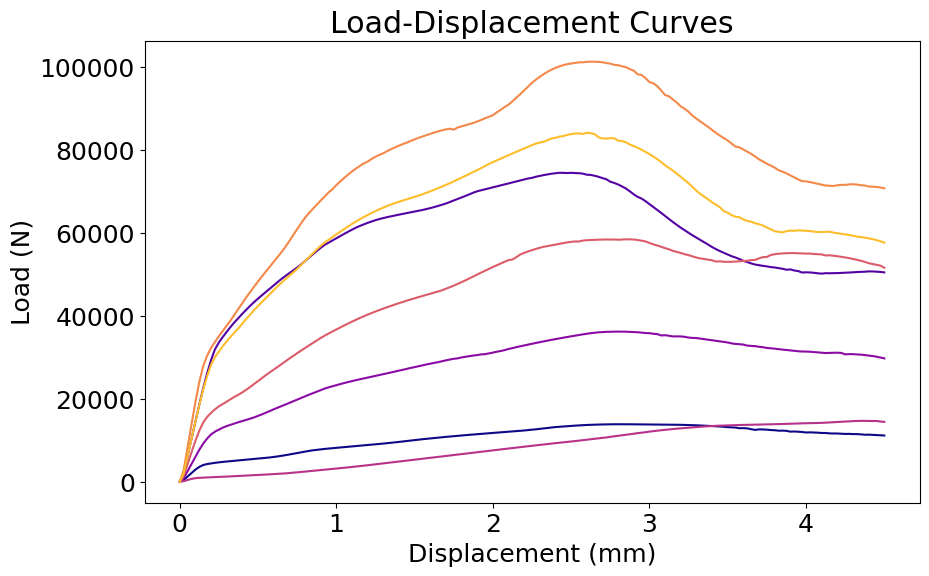}
		\end{minipage}
		\label{fig: load-displacement of training samples}
        }
        \subfigure[Stress-Strain Curves of Training Samples]{
        \begin{minipage}[b]{0.22\textwidth}
            \includegraphics[width=1\textwidth]{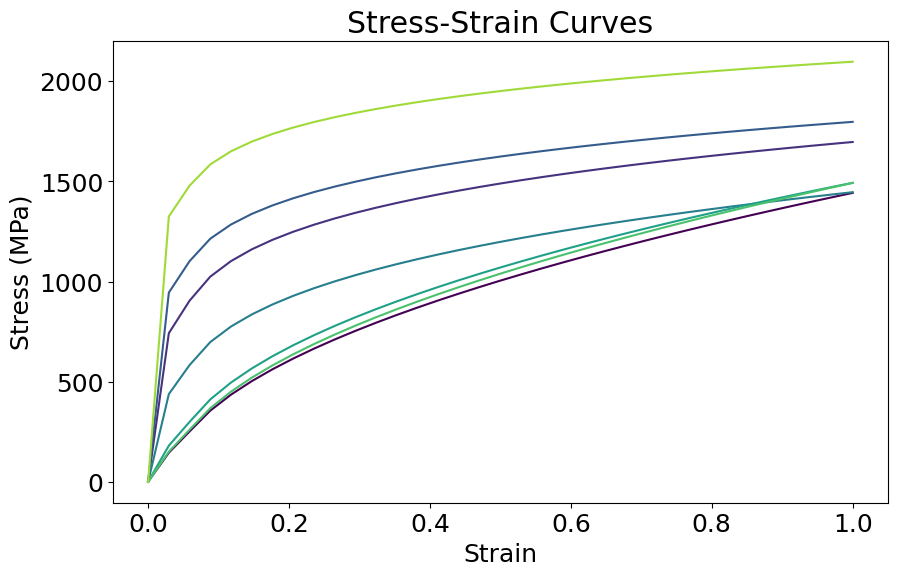}
        \end{minipage}
        \label{fig: Stress-Strain of training samples}
        }
	\caption{Load-Displacement Curves and Stress-Strain Curves of Training Samples}
	\label{fig2}
\end{figure}

\begin{figure}[!t]
	\centering
        \subfigure[Load-Displacement Curves of Testing Samples]{
		\begin{minipage}[b]{0.22\textwidth}
			\includegraphics[width=1\textwidth]{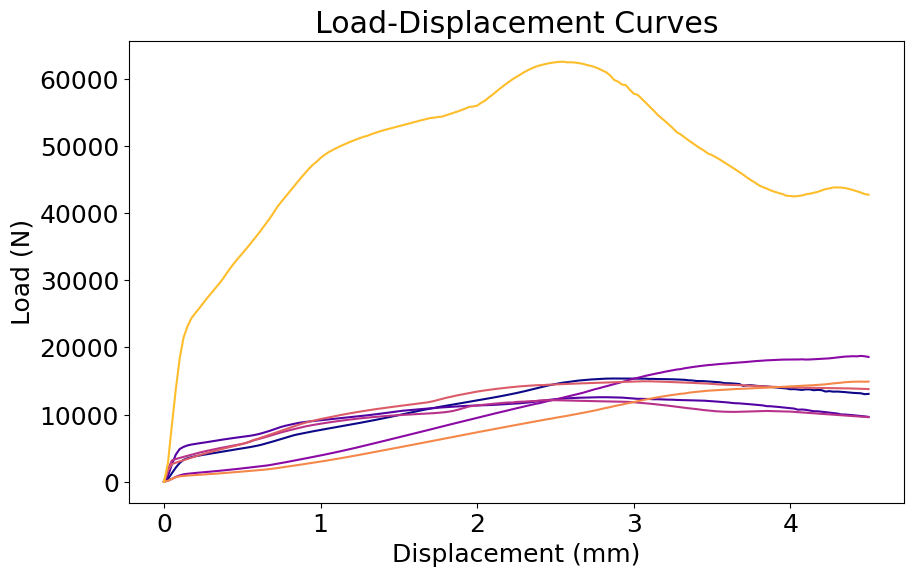}
		\end{minipage}
		\label{fig: load-displacement of testing samples}
        }
        \subfigure[Stress-Strain Curves of Testing Samples]{
        \begin{minipage}[b]{0.22\textwidth}
            \includegraphics[width=1\textwidth]{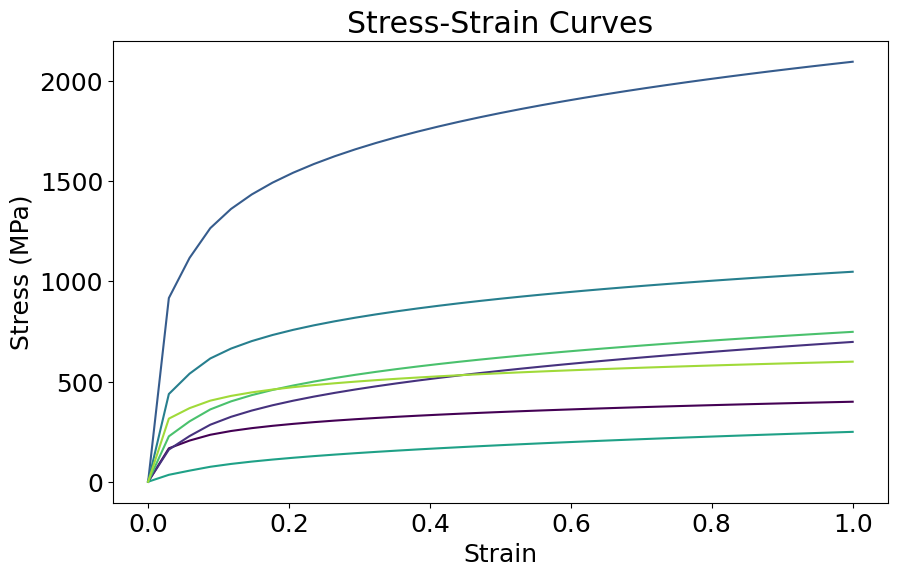}
        \end{minipage}
        \label{fig: Stress-Strain of testing samples}
        }
	\caption{Load-Displacement Curves and Stress-Strain Curves of Testing Samples}
	\label{fig3}
\end{figure}

This study employed a comprehensive dataset comprising 5,000 samples of high-strength steel to evaluate the proposed methodology. The dataset was divided into a training set of 4,500 samples and a testing set of 500 samples. Each sample exhibits variation in critical mechanical properties, including yield stress, which ranges from 20.98 MPa to 1907.53 MPa, strain hardening exponent, ranging from 0.068 to 0.4046, and thickness, spanning from 1 mm to 4 mm. These variations encapsulate a wide spectrum of material behaviors, ensuring the robustness and generalizability of the predictive model. The detailed characteristics of the dataset are presented in Table~\ref{Datasets Description}. Fig.~\ref{fig2} and Fig.~\ref{fig3} respectively display the SPT curves and stress-strain curves for the training and testing samples used in this study.

\begin{table}[b]
\centering
\caption{Datasets Description}
\label{Datasets Description}
\renewcommand{\arraystretch}{1.2} % 调整行间距
\begin{tabular}{@{}lcc@{}}
\toprule
\textbf{Attribute} & \textbf{Train} & \textbf{Test} \\ \midrule
\textbf{Number of Samples} & 4500 & 500 \\
\textbf{Young's Modulus (MPa)} & 70000 & 70000 \\
\textbf{Poisson's Ratio} & 0.35 & 0.35 \\
\textbf{Yield Stress (MPa)} & 20.98--1907.53 & 28.72--1823.16 \\
\textbf{Strain Hardening Exponent} & 0.068--0.4046 & 0.068--0.4046 \\
\textbf{Thickness (mm)} & 1--4 & 1--4 \\
\bottomrule
\end{tabular}
\end{table}

\begin{table}[b]
\centering
\caption{Experimental Results Comparison}
\label{Experimental Results Comparison}
\setlength{\tabcolsep}{5pt} % 调整列间距
\renewcommand{\arraystretch}{1.2} % 调整行间距
\begin{tabular}{@{}lcccc@{}}
\toprule
\textbf{Approaches} & \textbf{Max MAE} & \textbf{Min MAE} & \textbf{Max } $R^2$ & \textbf{Min } $R^2$ \\ \midrule
1D LSTM-based Model & \underline{17.44} & \underline{0.73} & \underline{0.999} & \underline{0.921} \\ 
ANN-based Model & 71.31 & 1.93 & 0.998 & 0.391 \\ 
GRU-based Model & 51.72 & 1.86 & 0.999 & 0.369 \\ 
Transformer-based Model & 450.77 & 6.25 & 0.999 & -0.631 \\ 
Proposed Model & \textbf{5.58} & \textbf{0.15} & \textbf{0.999} & \textbf{0.986} \\ 
\bottomrule
\end{tabular}
\end{table}

\begin{figure*}[htb]
	\centering
        \subfigure[Results Comparison of Sample 1]{
		\begin{minipage}[b]{0.45\textwidth}
			\includegraphics[width=1\textwidth]{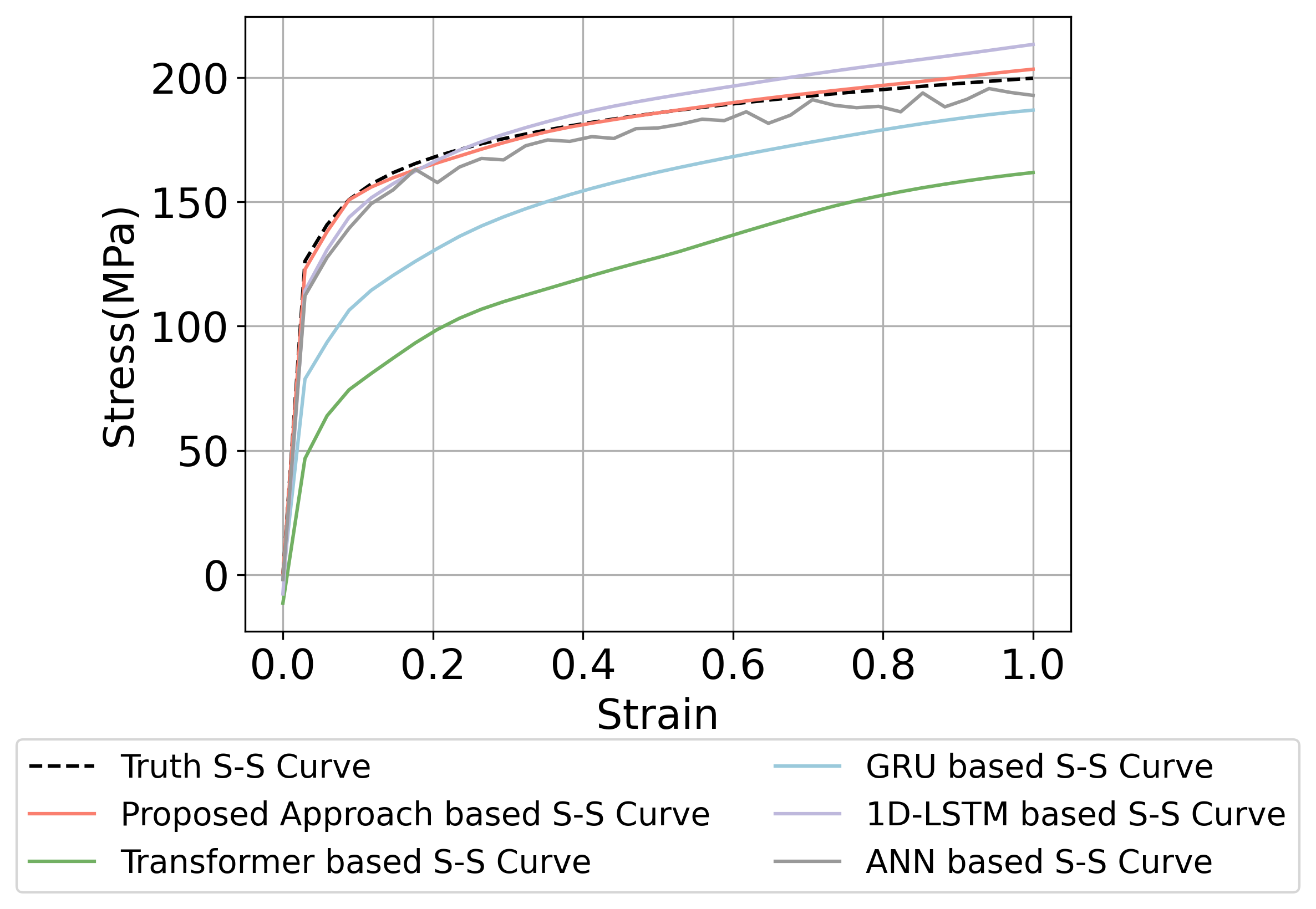}
		\end{minipage}
		\label{fig: Results comparison of Sample 1}
        }
        \subfigure[Results Comparison of Sample 2]{
        \begin{minipage}[b]{0.45\textwidth}
            \includegraphics[width=1\textwidth]{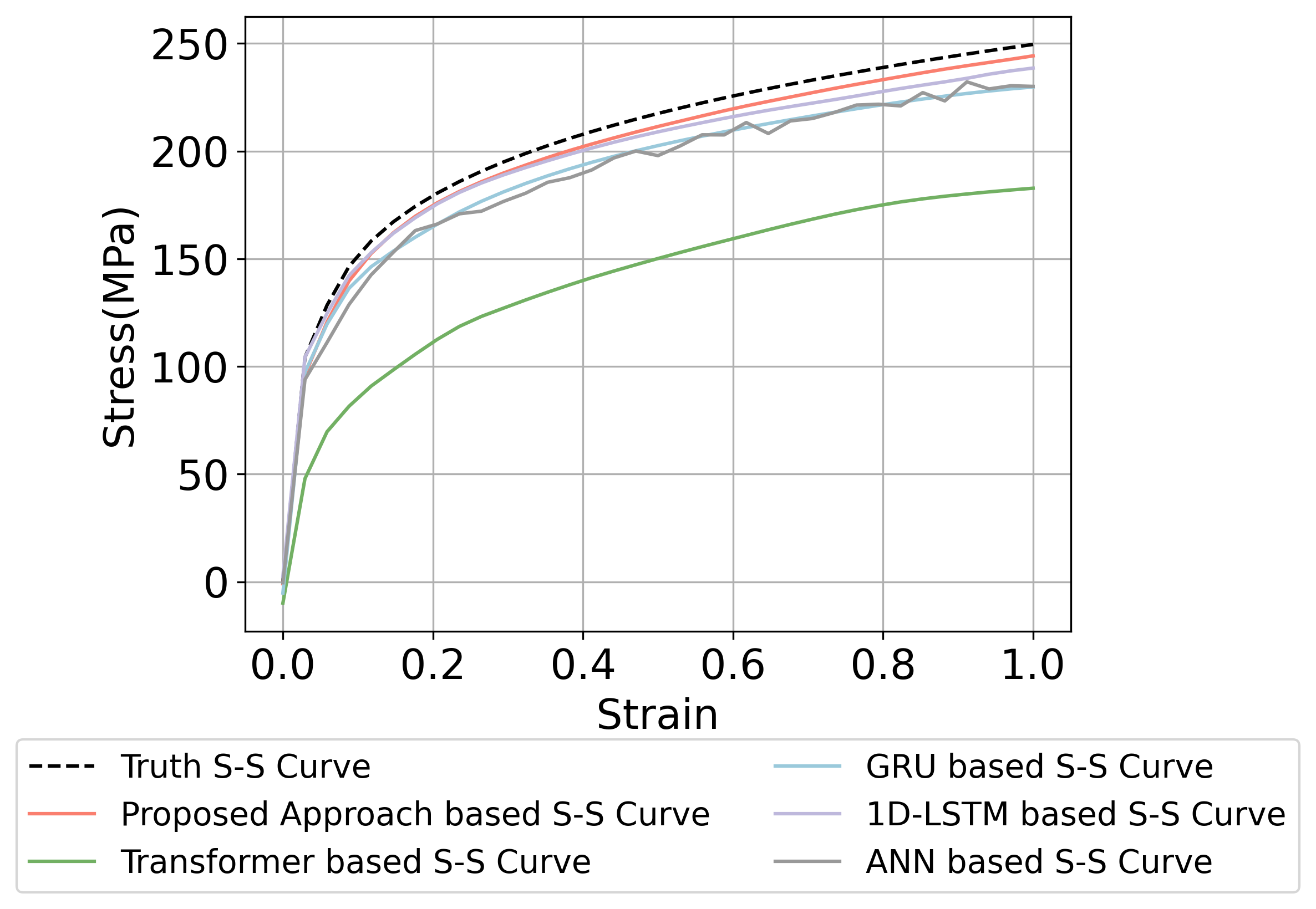}
        \end{minipage}
        \label{fig: Results comparison of Sample 2}
        }
        \subfigure[Results Comparison of Sample 3]{
        \begin{minipage}[b]{0.45\textwidth}
            \includegraphics[width=1\textwidth]{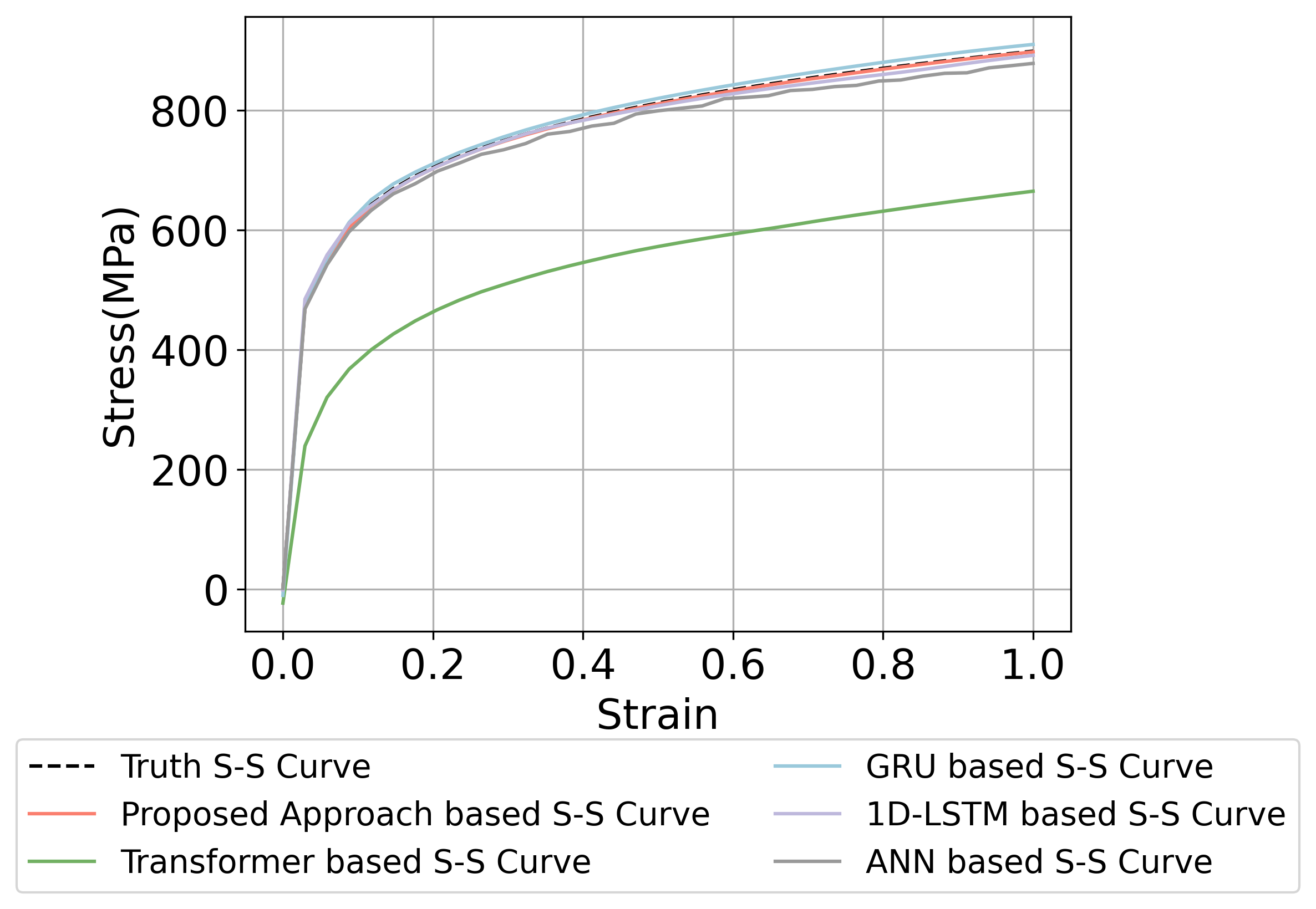}
        \end{minipage}
        \label{fig: Results comparison of Sample 3}
        }
        \subfigure[Results Comparison of Sample 4]{
        \begin{minipage}[b]{0.45\textwidth}
            \includegraphics[width=1\textwidth]{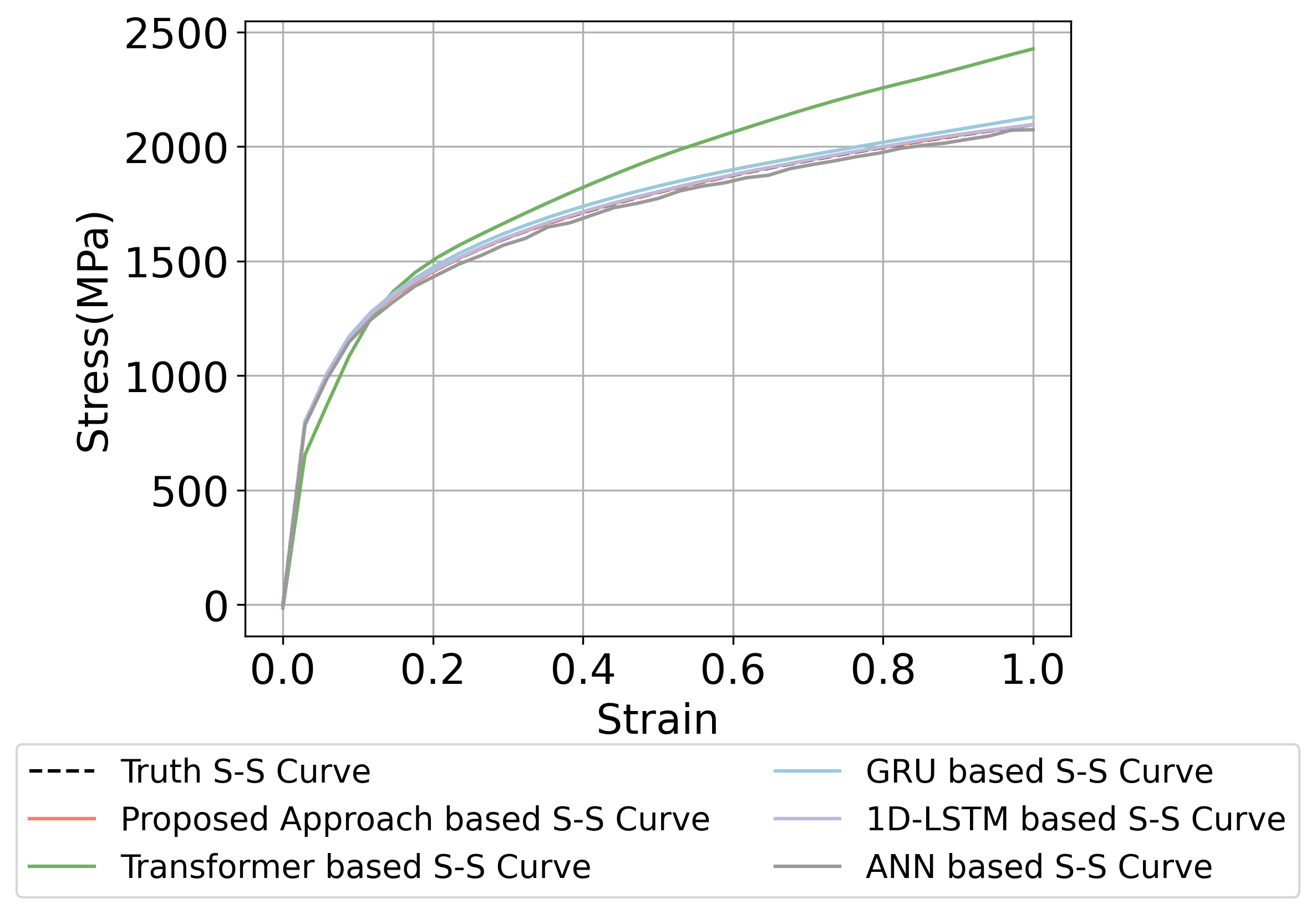}
        \end{minipage}
        \label{fig: Results comparison of Sample 4}
        }
	\caption{Results Comparison of Testing Samples}
	\label{fig4}
\end{figure*}

This study compares the proposed approach with a 1D LSTM-based model, ANN model \cite{song2020construction}, Gated Recurrent Unit (GRU)-based model \cite{mirzavand2023explainable}, and Transformer-based model \cite{10.5555/3305381.3305510}, evaluating various deep learning architectures for stress-strain curve prediction. Fig.~\ref{fig4} presents the comparison of experimental results.

In Fig.~\ref{fig4}, the four test samples were selected to encompass a broad range of stress levels, from approximately 200 MPa to 2500 MPa. Fig.~\ref{fig: Results comparison of Sample 1} illustrates that the prediction of ANN-based model exhibits significant fluctuations due to its inability to capture sequential dependencies, unlike Seq2Seq-based models. Regarding Fig.~\ref{fig: Results comparison of Sample 4}, the scale of the stress has been expanded to 2500 MPa, which diminishes the apparent fluctuations of the ANN model in the visualization. Table~\ref{Experimental Results Comparison} clearly demonstrates that the proposed approach outperforms the other four models in terms of accuracy. The maximum mean absolute error (MAE) for the proposed method is 5.58 MPa, and the minimum MAE is 0.15 MPa, and the corresponding R-squared statistic (R²) values are 0.999 and 0.986, respectively. The 1D LSTM-based model, which shares a similar structure with the proposed model but excludes the GAF transformation, achieves the second highest accuracy. For 1D LSTM-based model, the maximum and minimum MAE are 17.44 MPa and 0.73 MPa, respectively. While the 1D LSTM-based model performs well, the absence of the GAF algorithm results in a notable decrease in accuracy compared to the proposed approach, underscoring the effectiveness of the GAF-based feature extraction.

In contrast, the ANN-based model, which lacks a Seq2Seq structure, performs significantly worse, with a maximum MAE of 71.31 MPa and a minimum MAE of 1.93 MPa. Additionally, the predicted stress-strain curve generated by the ANN-based model lacks smoothness, suggesting that it struggles to capture the trend in the stress-strain relationship effectively. This result indicates that the ANN model is not well-suited for time-series data prediction in this context. For the GRU and Transformer-based models, which lack a cross-attention mechanism, the maximum MAE values are 51.72 MPa and 450.77 MPa, while the minimum MAE values are 1.86 MPa and 6.25 MPa, respectively. Notably, the Transformer-based model performs the poorest among all methods, likely due to insufficient training data, as transformer models typically require larger datasets to achieve optimal performance. Additionally, the higher complexity of the transformer model exacerbates overfitting when training data is limited, resulting in poor performance on the testing set, with \( R^2 \) value of -0.631. 

Overall, the results highlight the superiority of the proposed approach, particularly the effectiveness of the GAF transformation and the cross-attention mechanism in accurately predicting true stress-strain relationships in high-strength steel materials. Furthermore, the Seq2Seq model proves to be well-suited for predicting stress-strain sequence data from load-displacement sequence data.

\section{Conclusion and Future Work}
\label{Conclusion}
In conclusion, this study demonstrates an effective deep-learning methodology for predicting true stress-strain curves of high-strength steels using load-displacement data from SPT. The proposed model, which integrates a GAF-based transformation, 1D and 2D convolutional networks, and an LSTM-based Seq2Seq architecture with multi-head cross-attention, demonstrates superior accuracy and computational efficiency compared to both traditional methods and other deep learning approaches. The experimental results validate the model's capability to capture intricate dependencies within the data, underscoring the advantages of incorporating cross-attention mechanisms and spatial-temporal feature extraction for material property prediction.

Future research will extend the proposed model to various material types and testing conditions to evaluate its generalizability. Additionally, incorporating transfer learning algorithms for unlabeled materials could further broaden the applicability of the approach. Exploring lightweight models could enable real-time industrial applications for immediate material characterization.

\bibliographystyle{plainnat}
\bibliography{references}

\begin{thebibliography}{24}
\providecommand{\natexlab}[1]{#1}
\providecommand{\url}[1]{\texttt{#1}}
\expandafter\ifx\csname urlstyle\endcsname\relax
  \providecommand{\doi}[1]{doi: #1}\else
  \providecommand{\doi}{doi: \begingroup \urlstyle{rm}\Url}\fi

\bibitem[Cao et~al.(2022)Cao, Zu, Zhen, Li, and Wu]{cao2022determination}
Y.~Cao, Y.~Zu, Y.~Zhen, F.~Li, and G.~Wu.
\newblock Determination of the true stress-strain relations of high-grade pipeline steels based on small punch test correlation method.
\newblock \emph{International Journal of Pressure Vessels and Piping}, 199:\penalty0 104739, 2022.

\bibitem[Chen et~al.(2023)Chen, Yang, Zhong, Xi, and Liu]{10138915}
S.~Chen, R.~Yang, M.~Zhong, X.~Xi, and C.~Liu.
\newblock A random forest and model-based hybrid method of fault diagnosis for satellite attitude control systems.
\newblock \emph{IEEE Transactions on Instrumentation and Measurement}, 72:\penalty0 1--13, 2023.

\bibitem[Cornaggia et~al.(2018)Cornaggia, Cocchetti, Maier, and Buljak]{cornaggia2018inverse}
A.~Cornaggia, G.~Cocchetti, G.~Maier, and V.~Buljak.
\newblock Inverse structural analyses on small punch tests, with model reduction and stochastic approach.
\newblock In \emph{2018 IEEE International Conference on Environment and Electrical Engineering and 2018 IEEE Industrial and Commercial Power Systems Europe (EEEIC / I\&CPS Europe)}, pages 1--5. IEEE, 2018.

\bibitem[Deng et~al.(2024)Deng, Zhang, Dai, Shi, Zhong, and Li]{10637749}
H.~Deng, T.~Zhang, Y.~Dai, J.~Shi, Y.~Zhong, and H.~Li.
\newblock Deep non-rigid structure-from-motion: A sequence-to-sequence translation perspective.
\newblock \emph{IEEE Transactions on Pattern Analysis and Machine Intelligence}, 46\penalty0 (12):\penalty0 10814--10828, 2024.

\bibitem[Gehring et~al.(2017)Gehring, Auli, Grangier, Yarats, and Dauphin]{10.5555/3305381.3305510}
Jonas Gehring, Michael Auli, David Grangier, Denis Yarats, and Yann~N Dauphin.
\newblock Convolutional sequence to sequence learning.
\newblock In \emph{International conference on machine learning}, pages 1243--1252. PMLR, 2017.

\bibitem[Georgiev et~al.(2017)Georgiev, Dyer, and Black]{Georgiev2017Seq2SeqASR}
G.~Georgiev, C.~Dyer, and A.~W. Black.
\newblock Seq2seq models for end-to-end speech recognition.
\newblock In \emph{2017 IEEE Automatic Speech Recognition and Understanding Workshop (ASRU)}, pages 317--323. IEEE, 2017.

\bibitem[Gheini et~al.(2021)Gheini, Ren, and May]{gheini-etal-2021-cross}
Mozhdeh Gheini, Xiang Ren, and Jonathan May.
\newblock Cross-attention is all you need: Adapting pretrained transformers for machine translation.
\newblock \emph{2021 Conference on Empirical Methods in Natural Language Processing}, 11:\penalty0 1754--1765, 2021.

\bibitem[Hatakeyama-Sato(2021)]{HatakeyamaSato2021MachineLF}
K.~Hatakeyama-Sato.
\newblock Machine learning for material science.
\newblock \emph{The Brain \& Neural Networks}, 28\penalty0 (1):\penalty0 20--47, 2021.

\bibitem[Jan{\v{c}}a et~al.(2016)Jan{\v{c}}a, Siegl, and Hau{\v{s}}ild]{janvca2016small}
A.~Jan{\v{c}}a, J.~Siegl, and P.~Hau{\v{s}}ild.
\newblock Small punch test evaluation methods for material characterisation.
\newblock \emph{Journal of Nuclear Materials}, 481:\penalty0 201--213, 2016.

\bibitem[Kamaya and Lucas(2018)]{Kamaya2018SmallPunchTesting}
M.~Kamaya and G.~Lucas.
\newblock Small punch testing for structural materials: A review.
\newblock \emph{Journal of Nuclear Materials}, 510:\penalty0 255--269, 2018.

\bibitem[Klevtsov et~al.(2009)Klevtsov, Neshumaev, and Dedov]{klevtsov2009method}
I.~A. Klevtsov, D.~N. Neshumaev, and A.~V. Dedov.
\newblock A method of using miniature samples for determining mechanical properties of metal of power-generating equipment at thermal power stations in estonia.
\newblock \emph{Thermal Engineering}, 56:\penalty0 426--431, 2009.

\bibitem[Li et~al.(2018)Li, Peng, and Zhou]{li2018construction}
K.~Li, J.~Peng, and C.~Zhou.
\newblock Construction of whole stress-strain curve by small punch test and inverse finite element.
\newblock \emph{Results in Physics}, 11:\penalty0 440--448, 2018.

\bibitem[Mirzavand~Borujeni et~al.(2023)Mirzavand~Borujeni, Arras, Srinivasan, and Samek]{mirzavand2023explainable}
Sara Mirzavand~Borujeni, Leila Arras, Vignesh Srinivasan, and Wojciech Samek.
\newblock Explainable sequence-to-sequence gru neural network for pollution forecasting.
\newblock \emph{Scientific Reports}, 13\penalty0 (1):\penalty0 9940, 2023.

\bibitem[Song et~al.(2020)Song, Li, Cao, and Ma]{song2020construction}
M.~Song, X.~Li, Y.~Cao, and S.~Ma.
\newblock Construction of true stress-strain curve of metallic material by artificial neural network and small punch test.
\newblock \emph{Journal of Physics: Conference Series}, 1676\penalty0 (1):\penalty0 012130, 2020.

\bibitem[Sterjovski et~al.(2005)Sterjovski, Nolan, Carpenter, Dunne, and Norrish]{STERJOVSKI2005536}
Z.~Sterjovski, D.~Nolan, K.~R. Carpenter, D.~P. Dunne, and J.~Norrish.
\newblock Artificial neural networks for modelling the mechanical properties of steels in various applications.
\newblock \emph{Journal of Materials Processing Technology}, 170\penalty0 (3):\penalty0 536--544, 2005.

\bibitem[Sutskever et~al.(2014)Sutskever, Vinyals, and Le]{Sutskever2014SequenceToSequence}
I.~Sutskever, O.~Vinyals, and Q.~V. Le.
\newblock Sequence to sequence learning with neural networks.
\newblock \emph{Advances in Neural Information Processing Systems}, 3\penalty0 (7):\penalty0 3104--3112, 2014.

\bibitem[Wang et~al.(2023)Wang, Zhang, Zhou, Xue, Jia, and Zhu]{wang2023prediction}
H.~Wang, C.~Zhang, B.~Zhou, S.~Xue, P.~Jia, and X.~Zhu.
\newblock Prediction of triaxial mechanical properties of rocks based on mesoscopic finite element numerical simulation and multi-objective machine learning.
\newblock \emph{Journal of King Saud University-Science}, 35\penalty0 (7):\penalty0 102846, 2023.

\bibitem[Wang and Wang(2015)]{WANG201568}
Jie Wang and Jun Wang.
\newblock Forecasting stock market indexes using principle component analysis and stochastic time effective neural networks.
\newblock \emph{Neurocomputing}, 156:\penalty0 68--78, 2015.

\bibitem[Xue et~al.(2023)Xue, Yang, Chen, Tian, and Wang]{10193811}
Y.~Xue, R.~Yang, X.~Chen, Z.~Tian, and Z.~Wang.
\newblock A novel local binary temporal convolutional neural network for bearing fault diagnosis.
\newblock \emph{IEEE Transactions on Instrumentation and Measurement}, 72:\penalty0 1--13, 2023.

\bibitem[Yang et~al.(2021)Yang, Yang, and Huang]{s21237894}
Z.~Yang, R.~Yang, and M.~Huang.
\newblock Rolling bearing incipient fault diagnosis method based on improved transfer learning with hybrid feature extraction.
\newblock \emph{Sensors}, 21\penalty0 (23):\penalty0 7894, 2021.

\bibitem[Yang et~al.(2025)Yang, Zou, Huang, Yang, Zhang, Tong, Kong, Zhan, and Liu]{yang2025machine}
Z.~Yang, J.~Zou, L.~Huang, R.~Yang, J.~Zhang, C.~Tong, J.~Kong, Z.~Zhan, and Q.~Liu.
\newblock Machine learning-based extraction of mechanical properties from multi-fidelity small punch test data.
\newblock \emph{Advances in Manufacturing}, 2195\penalty0 (3597):\penalty0 1--14, 2025.

\bibitem[Zhan and Li(2021)]{zhan2021machine}
Z.~Zhan and H.~Li.
\newblock Machine learning based fatigue life prediction with effects of additive manufacturing process parameters for printed ss 316l.
\newblock \emph{International Journal of Fatigue}, 142:\penalty0 105941, 2021.

\bibitem[Zhang and Subic(2019)]{Zhang2019AdvancedMaterials}
D.~Zhang and A.~Subic.
\newblock Advanced materials and their applications in small punch testing: A review.
\newblock \emph{Materials \& Design}, 171:\penalty0 107734, 2019.

\bibitem[Zhang et~al.(2023)Zhang, Yang, Yue, Lim, and Wang]{10198368}
Y.~Zhang, R.~Yang, Y.~Yue, E.~G. Lim, and Z.~Wang.
\newblock An overview of algorithms for contactless cardiac feature extraction from radar signals: Advances and challenges.
\newblock \emph{IEEE Transactions on Instrumentation and Measurement}, 72:\penalty0 1--20, 2023.

\end{thebibliography}

\end{document}